\documentclass[conference]{IEEEtran}
\IEEEoverridecommandlockouts
\usepackage{cite}
\usepackage{amsmath,amssymb,amsfonts}
\usepackage{graphicx}
\usepackage{textcomp}
\usepackage{xcolor}
\def\BibTeX{{\rm B\kern-.05em{\sc i\kern-.025em b}\kern-.08em
    T\kern-.1667em\lower.7ex\hbox{E}\kern-.125emX}}
    \usepackage{algorithm}
\usepackage{algpseudocode}
\usepackage{amsmath}
\newcommand{\modelname}{3DBonsai}

\usepackage{graphicx}
\usepackage[utf8]{inputenc} % allow utf-8 input
\usepackage[T1]{fontenc}    % use 8-bit T1 fonts
\usepackage{hyperref}       % hyperlinks
\usepackage{url}            % simple URL typesetting
\usepackage{booktabs}       % professional-quality tables
\usepackage{amsfonts}       % blackboard math symbols
\usepackage{nicefrac}       % compact symbols for 1/2, etc.
\usepackage{microtype}      % microtypography
\usepackage{xcolor}         % colors
\usepackage{algorithm}
\usepackage{algpseudocode}
\usepackage{amsmath}
\usepackage{comment}
\usepackage{footmisc}   % 增强脚注控制（推荐）
\begin{document}

\title{3DBonsai: Structure-Aware Bonsai Modeling Using Conditioned 3D Gaussian Splatting}

% \author{Anonymous ICME submission}
\author{
Hao Wu, Hao Wang\textsuperscript{\textdagger}, Ruochong Li, Xuran Ma, Hui Xiong\textsuperscript{\textdagger}\thanks{\textdagger Corresponding authors.}\\
The Hong Kong University of Science and Technology (Guangzhou), China
}

\maketitle
\footnotetext{This work was supported in part by the National Natural Science Foundation of China(Grant No.92370204), in part by the National Key R\&D Program of China(Grant No.2023YFF0725001), in part by Guangzhou-HKUST(GZ) Joint Funding Program(Grant No.2023A03J0008), in part by the Education Bureau of Guangzhou Municipality.}

\begin{abstract}
Recent advancements in text-to-3D generation have shown remarkable results by leveraging 3D priors in combination with 2D diffusion. 
However, previous methods utilize 3D priors that lack detailed and complex structural information, limiting them to generating simple objects and presenting challenges for creating intricate structures such as bonsai.  
In this paper, we propose \modelname, a novel text-to-3D framework for generating 3D bonsai with complex structures. Technically, we first design a trainable 3D space colonization algorithm to produce bonsai structures, which are then enhanced through random sampling and point cloud augmentation to serve as the 3D Gaussian priors. 
We introduce two bonsai generation pipelines with distinct structural levels: fine structure conditioned generation, which initializes 3D Gaussians using a 3D structure prior to produce detailed and complex bonsai, and coarse structure conditioned generation, which employs a multi-view structure consistency module to align 2D and 3D structures.
Moreover, we have compiled a unified 2D and 3D Chinese-style bonsai dataset. Our experimental results demonstrate that \modelname~significantly outperforms existing methods, providing a new benchmark for structure-aware 3D bonsai generation. Demos and more details are available at \href{https://3dbonsai.github.io/}{https://3dbonsai.github.io/}.
\end{abstract}


\section{Introduction}
3D assets are significant in various industrial sectors, such as game development, film production, metaverse, etc. Creating 3D assets is challenging since it demands highly professional and extensive design experience. Recently, text-to-3D object generation methods~\cite{wang2024prolificdreamer,xu2023-dream3d,lin2023magic3d,tang2023-dreamgaussian,yi2023-gaussiandreamer} have shown notable capabilities in producing 3D assets, including the common objects in Objaverse dataset~\cite{deitke2023objaverse}. 
% Additionally, some works~\cite{tang2023-dreamgaussian,gupta20233dgen} have integrated 3D priors with 2D diffusion using 3D Gaussian Splatting (3DGS) to generate higher-quality 3D objects. 
However, relevant studies on generating 3D objects with complex and irregular structures are still rare.

% This issue is particularly evident in objects with irregular structures, such as bonsai. 
% these approaches struggle with complex objects.
% It is observed the generation of 3D priors and the 2D diffusion process are hard to accommodate complex structures effectively. This issue is particularly evident in objects with complex and irregular structures, such as bonsai.

% 3D assets hold an unparalleled significance in various industrial sectors such as game development, film production, and the metaverse. Due to the complex production process of 3D assets, which demands high professionalism and extensive design experience from designers, the generation of 3D assets becomes even more meaningful. Recently, 2D diffusion models have demonstrated unprecedented excellence in image generation, producing images with high generalization capabilities and fidelity, as well as exceptional style transfer abilities. A series of efforts aim to leverage the potent potential of 2D diffusion models to address 3D issues, such as text-to-3D and img-to-3D. However, the outcomes for complex objects like trees and vegetation in 3D generation have been less than ideal, often encountering problems such as distortion in generated results, multi-face issues, and slow generation speeds.

This paper investigates an open research problem of generating 3D bonsai. Bonsai refers to the traditional Asian art of growing and shaping miniature trees in containers. This is a challenging yet valuable task, due to the structure complexity and aesthetic standard of target 3D bonsai.
Previous methods for generating 3D complex objects can be divided into  
% is inherently limited due to their complexity and the challenge of maintaining 3D consistency. 
% This paper takes the generation of complex plants as an intriguing example, hoping to propose a paradigm for the generation of complex objects.
% Currently, there are 
two principal categories. The first involves the optimization-based 2D lifting methods~\cite{poole2022dreamfusion,tang2023-dreamgaussian}. 
They introduce Score Distillation Sampling (SDS)~\cite{poole2022dreamfusion} to guide the synthesis of 3D objects through 2D diffusion.
This approach mitigates the lack of 3D or multi-view data for training 3D models. However, SDS loss can lead to oversaturated and oversmoothed results. This limitation makes it difficult for the model to effectively accommodate objects with complex and irregular structures, such as bonsai.

The second category of method adopts procedural generation~\cite{raistrick2023-infinigen,sun2023-3dgpt}. 
% In some works on 3D scene generation, procedural tree generation programs are common. 
These methods leverage programming capabilities in 3D modeling software, such as Blender, to generate the 3D assets automatically. 
% They can produce decent results but are limited by generating only a few styles or having low style transferability. 
To be specific, Infinigen~\cite{raistrick2023-infinigen} attempted to use procedural code to generate trees but was limited by the defined tree species from blender and generation algorithms, resulting in somewhat monotonous outcomes. 3DGPT \cite{sun2023-3dgpt} attempted to combine Large Language Models (LLMs) with Blender for the design of 3D assets, yet still encountered issues with fidelity and artistic style transfer in the generation of complex objects.

% To be specific, xxx
 
% Infinigen~\cite{raistrick2023-infinigen} attempted to use procedural code to generate trees but was limited by the defined tree species from blender and generation algorithms, resulting in somewhat monotonous outcomes. 3DGPT\cite{sun2023-3dgpt} attempted to combine Large Language Models (LLMs) with Blender for the design of 3D assets, yet still encountered issues with fidelity and artistic style transfer in the generation of complex objects.
% the 2D diffusion and 3D generation process of the above approaches can hardly accommodate objects with complex and irregular structures effectively, such as bonsai. 
% Typically, structures in previous methods guided by either 2D or 3D diffusion tend to be simplistic, making it challenging to generate complex objects with detailed structures.

To solve these problems in complex 3D object generation, we propose the \modelname~framework, designed to generate complex 3D bonsai effectively and efficiently.  
Technically, our approach combines 3D structure priors with structure-aware 3D Gaussian splatting. Firstly, we adapt the space colonization algorithm (SCA)~\cite{runions2007sca} to 3D space and generate initial 3D structure priors. To this end, we propose a novel 3D Space Colonization Algorithm (3D SCA), which is implemented through the generation of leaf nodes in 3D space and a trainable branch growth algorithm.
% Moreover, we collect a specialized bonsai dataset comprising RGB and depth map, which is employed to train the 3D SCA to achieve stylized effects. 
Then, these structure priors generated by 3D SCA are further processed through two different structure-aware 3DGS. The first directly utilizes the 3D structure prior as an initial state for 3DGS, followed by a strong constraint applied to the bonsai structure generation.
Another approach employs a weaker constraint strategy, utilizing the 3D diffusion~\cite{jun2023shap-E,nichol2022pointE} to produce coarse 3D prior. 
Then the multi-view structure consistency learning is utilized to extract depth map information from the 3D structures across multiple views~\cite{zhang2023addingcontrolnet,ye2023ipa}. This process guides the splatting process by enhancing the complexity of 3D structures and improving the structure consistency across different viewpoints.
% While the initialization of 3DGS still derives from 3D diffusion, our structure merely serves as guidance to enrich the structural information generated. 
The above two pipelines transform the 3D structure prior to an initialized 3D Gaussian distribution, which is then refined using the 2D diffusion model.

We have compiled extensive datasets to support and validate our proposed method. We collect and release a dataset of Chinese-style 2D bonsai images and a smaller collection of high-quality 3D bonsai assets, which serve as the benchmark datasets and help with the bonsai generation quality assessment. 

Our proposed 3D bonsai generation framework demonstrates significant effectiveness in generating bonsai from complex structures. Our contributions are summarized as follows:

\begin{itemize}
 \item We propose a trainable 3D space colonization algorithm that serves as the structure guidance for structure-aware 3D Gaussian splatting, ensuring high fidelity of the generated bonsai while maintaining 3D consistency.
 \item Our framework introduces two pipelines for structure constraints within 3DGS: fine structure and coarse structure. We employ a 3D structure prior to detailed fine structure 3DGS and multi-view structure consistency learning module to achieve coarse structure 3DGS.
 \item We present a comprehensive bonsai dataset featuring 20 high-quality Chinese-style bonsai 3D models and over a thousand high-quality images.
 
\end{itemize}

\section{Related Work}
\vspace{-5pt}
\subsection{3D Gaussian Splatting}

Recently, the 3D Gaussian splatting (3DGS)~\cite{kerbl2023-3dgs} method has optimized most of the issues associated with the methods above through its efficient rendering approach and superior rendering quality. It stands out as an impressive and practical 3D representation method. Not only does it facilitate real-time rendering, but it can also be conveniently integrated into 3D asset creation pipelines. 
Several current methods utilize 3D Gaussian splatting to generate high-quality 3D assets. DreamGaussian~\cite{tang2023-dreamgaussian} employs a 3D Gaussian splatting framework using a single image to produce satisfactory outcomes. However, it faces common limitations like multi-face anomalies. Expanding on 3DGS, GaussianDreamer~\cite{yi2023-gaussiandreamer} and GSGEN~\cite{chen2024textto3d} address the text-to-3D task. They both leverage 3D priors for 3D consistency and use a 2D diffusion model for high-resolution detail refinement. Their approaches significantly boost the optimization speed and generation quality of the text-to-3D models. 

\subsection{3D Trees Models Generation}
The tree generation algorithms typically employ fractals and repetitive patterns, environmentally sensitive automata, and particle systems for the generation pipeline. L-systems~\cite{prusinkiewicz2012algorithmic}, grounded in fractal principles, utilize rule-based procedural generation to address the creation of various plant types. However, the tree models generated by L-systems still lack realism and artistic quality. Traditional sketch-based methods and data-driven methods are capable of accurately generating images or point clouds of plants based on user input.

Currently, neural network-based methods for generating detailed tree images have achieved impressive results~\cite{lee2023latentLsystem}, producing a wide variety of trees with high realism. However, common text-to-3D approaches for generating 3D models of trees often fall short, resulting in issues like blurriness, distortion, loss of detail, and structural confusion. Procedural 3D generation methods~\cite{raistrick2023-infinigen,sun2023-3dgpt}, known for producing high-precision and high-quality 3D assets, have gained widespread attention. Yet, they are limited by the variety of generated types and the simplicity of tree detail, constrained by program design. Most efforts are dedicated to generating 3D Scenes and also integrating Blender Python to produce 3D plant models.

\section{Method}

\begin{figure*}
\centering   
  \includegraphics[width=1\textwidth]{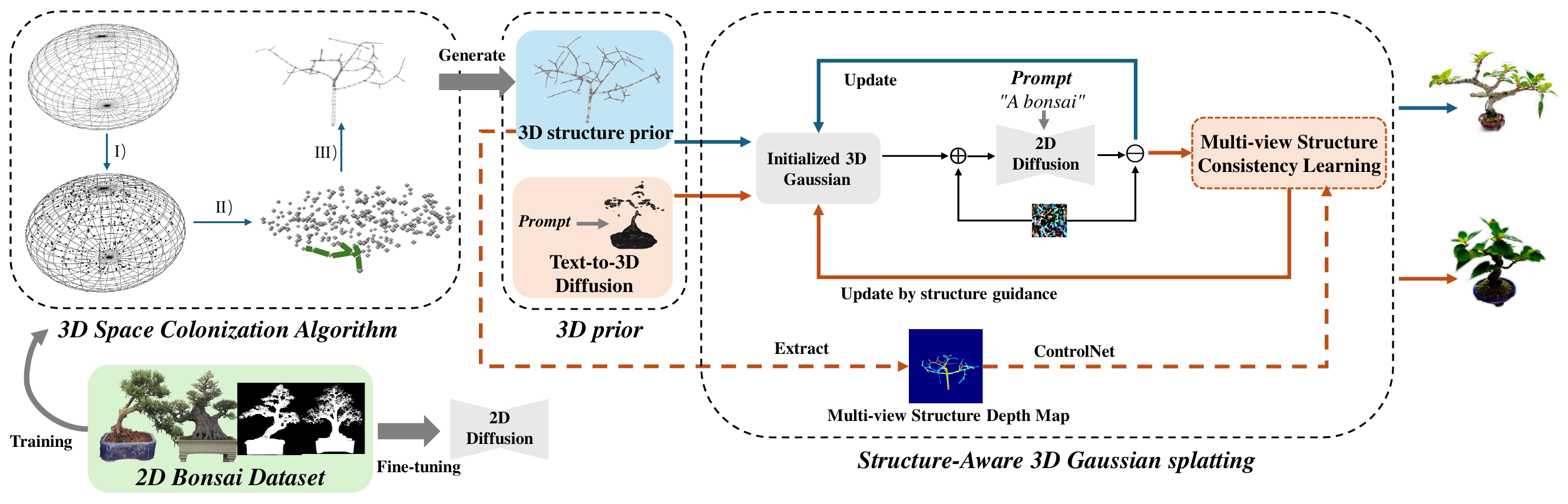}
  \vspace{-10pt}
\caption{\textbf{\modelname~Framework.} The 3D SCA module leverages our 2D Bonsai dataset to generate 3D structure priors. In structure-aware 3D Gaussian splatting, we implement two pipelines: (a) \textcolor{blue}{the fine structure (blue lines)} 3DGS uses the 3D structure from the 3D SCA as initialization to create 3D bonsai that closely align with the structure structure. (b) \textcolor{orange}{the coarse structure (orange lines)} 3DGS starts with the text-to-3D diffusion result as the initial state, followed by a 2D-3D structure consistency module to ensure 3D consistency by perceptually aligning the 3D structure prior.}
    \label{fig0:model}
    \vspace{-10pt}
\end{figure*}

\subsection{3D Space Colonization Algorithm}
Space Colonization Algorithm~\cite{runions2007sca} can be used in generating complex 2D structures, which is effective in procedural tree generation. Trees generated via SCA closely resemble real tree 2D structures.
% , making them suitable for various tree structure images. 
However, the original SCA fails to produce 3D structures directly.
% does not align with the characteristics of real-world 3D trees and fails to meet the requirements for 3D generation.
Therefore, we propose an improved trainable space colonization algorithm extended to 3D space. 
% ur framework utilizes it to generate the skeletal structure of bonsais as 3D prior information for the text-to-3D generation pipeline. 

\textbf{Leaf space definition.} We first generate a random distribution of points within a space to act as the initial leaves.
Initially, we delineate a spherical domain whose spatial constraints are designed to approximate the distribution of leaf nodes observed in real-world bonsai. This preliminary configuration ensures that the modeled space aligns with the natural bonsai structure.
\begin{equation}
\text{SPHERE\_DOMAIN} = \left\{
\begin{array}{ll}
x^2 + y^2 \leq R^2,  \\
z = R \pm \frac{\sqrt{R^2 - x^2 - y^2}}{2}
\end{array}
\right\},
\end{equation}
where $x^2 + y^2 \leq R^2$ specifies a circle with radius $R$ in the xy-plane, and $z$ represents an ellipsoid that is vertically centered at $R$ on the z-axis, with its vertical extent halved.
Subsequently, within this defined space, we randomly generate a specific number of points that act as the initial leaves, as depicted in Figure \ref{fig0:model} (\uppercase\expandafter{\romannumeral1}). The magnitude \(d\) is randomly selected within the interval \([0, R]\). Let \(P\) denote the random position of our leaf, defined as:
\begin{equation}
P = \vec{D} \cdot d,
\end{equation}

% After establishing the leaves, it is necessary to define the branches. 
\textbf{Branch growth learning.} Branches are segments that connect two points. To navigate structures from any branch, each branch must store references to its parent and its children, as well as its growth direction. 
% This storage facilitates frequent use of directional data. 
% To this end, we first define an attraction distance. 
As the structure grows, the branches are drawn towards previously defined leaves that are within the attraction distance. Figure \ref{fig0:model} (\uppercase\expandafter{\romannumeral2}) shows that the branches grow process in the 3D SCA. 
We determine the attraction points influencing each branch to enable our tree to branch out in multiple directions. If any attraction points are identified, a child branch is built, oriented towards the average direction of those points. Let \(B_{\alpha}\) and \(B_{\beta}\) define the branch \(\alpha\) and its child branch \(\beta\) respectively. 
We use \(\overrightarrow{D_{\beta}}\), \(pos_{\alpha}\), and  \(P_{\alpha}\) to represent the direction of the child of \(B_{\alpha}\), the position of the extremity of \(B_{\alpha}\), and the list of attraction points for \(B_{\alpha}\) with size \(i_{\alpha}\) respectively. The direction of the child branch is calculated by normalizing the sum of the vectors from the extremity of \(B_{\alpha}\) to each attraction point:
\begin{equation}
\overrightarrow{D_{\beta}} = \frac{\omega}{i_{\alpha}} \sum_{n=0}^{i_{\alpha}-1} \frac{ \overrightarrow{P_{\alpha}[n]} - \overrightarrow{pos_{\alpha}} }{ \| \overrightarrow{P_{\alpha}[n]} - \overrightarrow{pos_{\alpha}}\|},
\end{equation}
where $\omega$ is the trainable parameter with our proposed 2D bonsai dataset.
This formula is initially applied to the extremity of the 3D tree and will subsequently be applied to all branches, facilitating growth in diverse directions.

\noindent\textbf{Kill distance definition.} With the proposed branch growth algorithm, it is observed during growth, branches are attracted by many leaves, causing the branches to twist and deviate from their expected paths. To mitigate this issue, we define the kill distance, which is a threshold. If an attraction point is closer to a current branch than the kill distance, it is removed from the list of attraction points. Furthermore, it is notable the kill distance should be less than the attraction distance but greater than the predefined branch length. This adjustment helps prevent the undesired twisting of branches. 

\noindent \textbf{3D representation.}
After generating the branch graph for the 3D structure, it is necessary to endow it with a tangible volume within the 3D space. Consequently, the next step involves converting these branches into a mesh or point cloud to enable realistic and spatially coherent visualization and to facilitate further processing. To transform the branch structure into a mesh, we enhance our branch data structure to include the indices of vertices. For each branch, $S$ vertices are created around its endpoint, aligned along the branch direction. Faces are then generated from these vertices to form the mesh. Let $N_b$ and $N_v$ represent the number of branches and the total number of vertices in our mesh respectively, where $N_v = (N_b + 1) \cdot S$. Furthermore, the number of faces, $N_f$, is calculated as follows: $N_f = N_b \cdot S \cdot 2$.

Although we can obtain an initial mesh, it overlooks the fact that bonsai branches typically thicken as they approach the root. Branch extremities are added last to the branches list, allowing us to parse the branches from the end to the beginning and compute their sizes using an inverted growth model. We define \( r_e \) as the size of an extremity. If a branch has no children, it is an extremity: $s_b = r_e$. The size of a branch \( s_b \) is computed as follows:
\vspace{-5pt}
\begin{equation}
s_b = \left( \sum_{n=1}^{N_{ch}} \text{children}[n].\text{size}^{I_g} \right)^{\frac{1}{I_g}},
\end{equation}
\vspace{-8pt}

where \( N_{ch} \) is the number of branch children, \(\text{children}\) is the array of children of the branch, and \( I_g \) is the inverted growth factor. This approach ensures a natural gradation in branch thickness, replicating the authentic growth patterns observed in real bonsai.
Once the mesh is created, we can obtain a high-quality point cloud by conducting random sampling across the mesh at a specified density.

\subsection{Structure-Aware 3D Gaussian Splatting}
In this section, we present the structure-aware 3D Gaussian splatting through two distinct branches within this module, where we use structures and point clouds respectively as conditional input. 

% Each pipeline applies a different level of structure constraint, resulting in varied generated structures. However, both of them rely on the text-to-3D framework~\cite{tang2023-dreamgaussian,yi2023-gaussiandreamer} and the methods of 3DGS~\cite{kerbl2023-3dgs}. Below, we will detail the distinctions between these two approaches.

\subsubsection{Fine Structure Conditioned 3D Gaussian Splatting}
% In this branch, we use the structures generated by 3D SCA as conditions,
% The structure-anchor 3DGS is the pipeline with stronger constraints, implying that the structure of the final 3D object will closely align with our 3D structure prior. 
% which enables 3D bonsai generation with complex structures. 
% The reason for this effectiveness is that the structure's granularity produced by our 3D SCA is significantly finer than the priors generated by 3D diffusion methods such as Shap-E~\cite{jun2023shap-E}.

We utilize the 3D structure prior generated by 3D SCA to initialize the 3D structure. The 3D structure prior enables \modelname~ to generate plants with complex structures, such as bonsai. 

Following the conventional text-to-3D pipeline \cite{xu2023-dream3d, wang2024prolificdreamer, poole2022dreamfusion,chen2023fantasia3d,lin2023magic3d}, our model uses an SDS loss to integrate text into the 2D diffusion process during the splatting process. The formula for SDS loss is as follows:

\begin{equation}
\nabla_{\Theta} \mathcal{L}_{SDS} = \mathbb{E}_{t,p,\epsilon} \left[ w(t)\Delta Z  \frac{\partial I^{p}_{RGB}}{\partial \Theta} \right],
\end{equation}

where \( \mathbb{E}_{t,p,\epsilon} \) denotes the expectation over time \( t \), position \( p \), and noise \( \epsilon \). \(\Delta Z\) represents the expected difference in the diffusion process. \( I^{p}_{RGB} \) represents the RGB image rendered from the 3D Gaussians at position \( p \). \( \Theta \) represents all the optimizable parameters for the 3D Gaussians. \( w(t) \) is a weighting function that depends on time \( t \).The SDS loss~\cite{rombach2022highstablediffusion} aids in extending the 3D structure by incorporating corresponding semantic details from the input prompt. The training process translates the semantics of the text into multi-view images generated through the 2D bonsai diffusion model, which are subsequently mapped onto the 3D structure.

In summary, the fine structure 3D Gaussian splatting achieves more complex and clearer structures in 3D objects through the strong constraints imposed by the 3D structure prior.

% In summary, the fine structure 3D Gaussian splatting achieves further fine-tuning by maintaining the strong structural information of the generated model and adding textures relevant to the given prompt. This process is enhanced with additional 3D Gaussian adjustments, resulting in the rendering of our final output.
% \textbf{Text-to-3D.} The input for text-to-3D is a single text prompt. The SDS loss can be formulated as:

%specifically \( \Theta_i = \{ \mathbf{x}_i, s_i, \mathbf{q}_i, \alpha_i, \mathbf{c}_i \} \) for the \( i \)-th Gaussian. Here, \( \mathbf{x} \in \mathbb{R}^3 \) is the center of the Gaussian, \( s \in \mathbb{R}^3 \) is the scaling factor, \( \mathbf{q} \in \mathbb{R}^4 \) is the rotation quaternion, \( \alpha \in \mathbb{R} \) is the opacity value, and \( \mathbf{c} \in \mathbb{R}^3 \) is the color feature for volumetric rendering.

\subsubsection{Coarse Structure Conditioned 3D Gaussian Splatting}
% The structure-guidance 3DGS offers a feasible pipeline for using the 3D structure prior as guidance in the multi-view 2D structure consistency. 
Since fine structure 3DGS imposes strict structure constraints on the 3D Gaussian generation, we propose another pipeline, coarse structure 3DGS, which is designed to allow more freedom during structure-conditioned bonsai generation.
% learn the complexity of the structure without directly imposing constraints. 
% Consequently, we initially follow the setting similar to that used in GaussianDreamer~\cite{yi2023-gaussiandreamer}; 
Technically, based on the input prompt, the coarse structure 3DGS first utilizes 3D diffusion~\cite{jun2023shap-E} to generate point clouds as the initialized 3D Gaussians. The generated point clouds facilitate the maintenance of multi-view consistency in 2D diffusion and provide a rough outline of the bonsai based on text prompts.
% Although this constraint is not as precise or high-quality, it serves as effective guidance.

\textbf{Multi-view structure consistency learning.} Effectively integrating the 3D prior structure into the multi-view 2D features for constraints remains a challenge, as Stable Diffusion \cite{rombach2022highstablediffusion} generates multi-view 2D images during the splatting process.
To achieve this, we dynamically extract multi-view depth maps from the 3D structure prior, translating structure information into multi-view images. This process builds on the method used by the text-compatible image prompt adapter~\cite{zeng2023ipdreamer,ye2023ipa}, equipped with a pre-trained depth map ControlNet~\cite{zhang2023addingcontrolnet} to accurately interpret the depth map data. Then, the coarse structure 3DGS employs two decoupled cross-attention mechanisms to integrate additional structure information of the depth map with the original image features.
% formula

Similar to the previously mentioned text-to-3D pipeline~\cite{poole2022dreamfusion, yi2023-gaussiandreamer}, we also applied a 3D generation framework based on the SDS loss~\cite{poole2022dreamfusion}. During training, we optimize the following loss to ensure multi-view structure consistency while keeping the parameters of the pre-trained diffusion model frozen: 
% This mechanism is trained on a dataset containing depth map-image pairs, using the same training objective as the original stable diffusion model.
\begin{equation}
L_{\text{s}} = \mathbb{E}_{x_0, \epsilon, c_i, c_{cd}, t} \left\| \epsilon - \epsilon_\theta (x_t, c_i, c_{cd}, t) \right\|^2,
\end{equation}
where \(x_0\) denotes the real data with conditions. \( t \in [0, T] \) is the time step of the diffusion process, \(c_i, c_{cd}\) denotes the condition of the depth map processed by ControlNet~\cite{zhang2023addingcontrolnet} and the condition of an image. $\epsilon$ represents the training objective of diffusion.  We also randomly drop image conditions in the training stage to enable structure-free guidance in the inference stage:
\begin{equation}
\hat{\epsilon}_\theta (x_t, c_i, t) = \alpha\epsilon_\theta (x_t, c_i, t) + (1 - \alpha) \epsilon_\theta (x_t, t),
\end{equation}

When $\alpha$ is zeroed out, the model becomes the original SD image-to-image model. Additionally, in this process, we utilize the information from the depth maps and images to jointly maintain 3D consistency. 
Inspired by the concept of decoupled cross-attention in the text-compatible image prompt adapter~\cite{ye2023ipa}, we employed two cross-attention mechanisms. The first cross-attention mechanism extracts features from the depth map image, and to maintain maximal 3D consistency, this mechanism remains immutable. The second cross-attention mechanism extracts features from images generated by the 2D diffusion model. We define the final formula for the decoupled cross-attention as follows, where $\beta$ controls the influence of features from the images generated by the 2D diffusion model on the 3D Gaussians:

\vspace{-10pt}
\begin{equation}
\mathbf{Z}^{\text{c}} = \text{Attention}(\mathbf{Q}, \mathbf{K}, \mathbf{V}) + \beta \cdot \text{Attention}(\mathbf{Q}, \mathbf{K}', \mathbf{V}').
\end{equation}

where $\mathbf{Q}$ is the query from depth map features. $\mathbf{K}, \mathbf{V}$ are key and value matrices from depth map features. $\mathbf{K}'$, $\mathbf{V}'$ are trainable key and value matrices from the image features.  \( \beta \) is the weight parameter. When \( \beta = 0 \), the decoupled attention becomes the cross-attention for the depth-to-image feature.

It is worth noting that our coarse structure 3DGS pipeline embeds multi-view information from the depth map into the splatting process, ensuring a consistent and detailed 3D representation. This process effectively optimizes 3D consistency, enhancing the fidelity and structural accuracy of the generated 3D bonsai.
% By integrating features across various views, we maintain structure consistency in 2D and effectively incorporate 3D structure cues into the 2D diffusion process to guide 3D reconstruction.

\begin{figure*}
\centering   
  \includegraphics[width=0.96\textwidth]{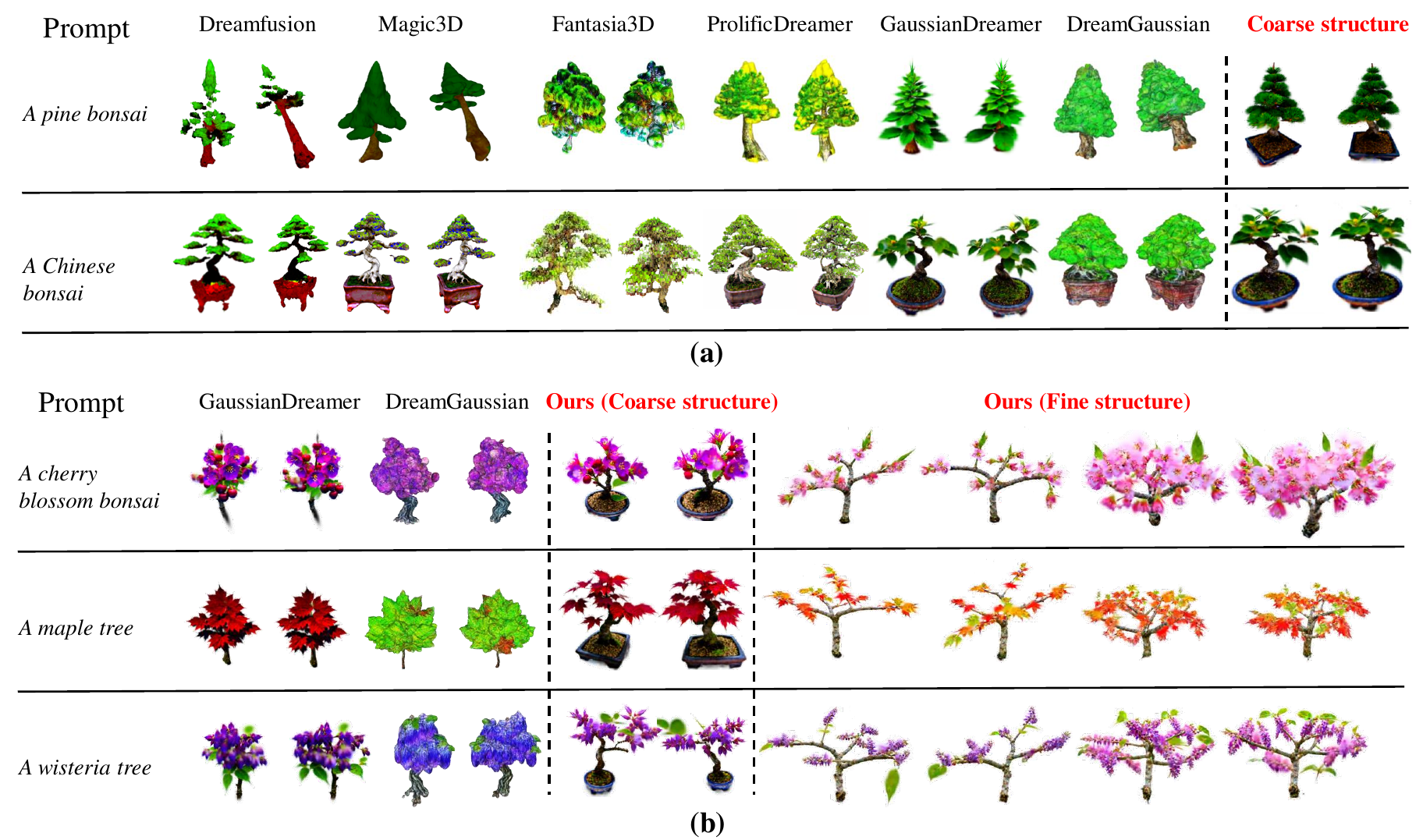}
  \vspace{-10pt}
\caption{\textbf{Qualitative comparisons} (a) Qualitative comparisons for image-to-3D between our \modelname(coarse structure)~and DreamFusion~\cite{poole2022dreamfusion}, Magic3D~\cite{lin2023magic3d}, Fantasia3D~\cite{chen2023fantasia3d}, ProlificDreamer~\cite{wang2024prolificdreamer}, GaussianDreamer~\cite{yi2023-gaussiandreamer}, DreamGaussian~\cite{tang2023-dreamgaussian}. The comparison results include the two best baselines and our generation results through fine and coarse structure pipelines; (b) Qualitative comparisons between two pipelines of \modelname~and GaussianDreamer~\cite{yi2023-gaussiandreamer}, DreamGaussian~\cite{tang2023-dreamgaussian}.}
    \label{fig1:result}
    \vspace{-10pt}
\end{figure*}

\section{Experiments}
\subsection{Implementation Details} %modified
% Our \modelname train 1200 step in 
The \text{\modelname} employs two distinct initializations for 3D Gaussian splatting: a structure point cloud generated by our trainable 3D SCA for the fine structure 3DGS, and the initialized 3D Gaussians generated by Shap-E~\cite{jun2023shap-E} for the coarse structure 3DGS. In 3DGS, we utilize the Stable Diffusion 2.1 model~\cite{rombach2022highstablediffusion} for generating multi-view images via 1200 steps in the splatting process. 
We employ four strategically chosen views with a camera distance range from 2.5 to 4 for the splatting settings, while the camera angles are randomly chosen to maximize view diversity. 
The training extends over 1200 steps, with an evaluation phase every 200 steps to ensure optimal parameter adjustments and model performance. The resolution for the Stable Diffusion images is set at 1024×1024. 
The text guidance in Stable Diffusion is achieved through SDS Loss, ensuring fine-grained fidelity in texture and depth. 
Additionally, we utilize the pre-trained IP-Adapter\_sd15 to seamlessly integrate depth and textual cues into the 3D generation process. Our model employs the Adam optimizer, configured with a learning rate (lr) of 0.001, betas set to [0.9, 0.99] for momentum parameters, and an epsilon (eps) value of \(1 \times 10^{-15}\) to improve numerical stability.
All experiments are performed and measured with an NVIDIA A6000 (48G) GPU. 

\subsection{Dataset} %modified
While the traditional SCA inherently produces a random distribution of point clouds,  our 3D SCA modifies this process to better replicate the real-world characteristics of bonsai structures.
Consequently, we have compiled a diverse dataset of 2D bonsai images to train our 3D SCA. This training aims to enable the algorithm to create bonsai structures that embody the characteristics of Chinese art. 
Our dataset includes 1233 bonsai images, and we also provide their corresponding masks to facilitate model training. 
Additionally, we offer 20 high-quality 3D bonsai mesh models for quality assessment and to aid future research, extracted using high-quality video footage and the Luma AI techniques.

% Some details about the framework
\subsection{Qualitative Evaluation}
We compare our methods with some of the latest text-to-3D approaches~\cite{poole2022dreamfusion,lin2023magic3d,chen2023fantasia3d,wang2024prolificdreamer,yi2023-gaussiandreamer,tang2023-dreamgaussian}, as illustrated in Figure \ref{fig1:result}. For the text-to-3D evaluations, we carefully select two views of suitable images generated by each 3D model to ensure a fair comparison. 
In Figure \ref{fig1:result}(a), We have compared five models with our coarse structure 3DGS using two different prompts, only GaussianDreamer~\cite{yi2023-gaussiandreamer} approaches our results. The other models perform significantly worse, and our model displays richer details than the rest. In Figure \ref{fig1:result}(b), we have selected the two better baselines to compare our coarse structure and fine structure 3DGS across more cases. Both of our pipelines offer superior generation quality, detail, and structural control compared to these baselines. These qualitative results demonstrate the effectiveness of our 3D structure prior within the framework.
\vspace{-10pt}
\subsection{Quantitative Evaluation}
Recently, no definitive metric exists for evaluating text-to-3D generation due to the complexity of the task and the open-domain nature of the generated objects, making quantitative assessments particularly challenging. 
Nevertheless, we align with the practices of quantitative evaluation in text-to-3D works by utilizing CLIP-based metrics for our assessments.
Specifically, we measure the average CLIP score between text and 3D renderings using variants of the CLIP~\cite{radford2021clip} model, CLIP ViT-bigG-14 and CLIP ViT-L/14. The evaluation is conducted with 30 prompts, each rendered from 120 viewpoints of the corresponding 3D outputs. 
According to the quantitative results in Table~\ref{tab0:quantitative}, our framework outperforms other text-to-3D models. 
However, the CLIP score primarily measures the relevance between text and image pairs, showing that our framework excels in this aspect compared to other models. Nonetheless, it does not capture the advantages of our model in terms of complexity and diversity.
Therefore, we conduct a user study comparing it with GaussianDreamer~\cite{yi2023-gaussiandreamer}, with approximately 68\% of participants expressing a preference for the results produced by our method over the baseline.

Our trainable 3D SCA demonstrates significant improvement over the untrained random 3D SCA. We produce 2D images from four fixed views of both sets of 3D structure priors and calculate the Frechet Inception Distance (FID)~\cite{heusel2017gans} against our bonsai dataset. The comparison of average values shows that after employing the trainable 3D SCA, the FID performance improved from 138 to 74.

% \begin{table}
% \centering
% % \renewcommand{\tabcolsep}{3mm}
% \begin{tabular}{l|ccc}
% \toprule
% \textbf{Method} &  SSIM & LPIPS & FID \\\midrule
% SCA (Random) & 0.40 & 2.37 & 1.35 \\
% SCA (Pre-trained) & 2.93 & 9.23 & 6.05\\\midrule
% \modelname & \textbf{5.64} & \textbf{18.50} & \textbf{12.09}\\
% \bottomrule
% \end{tabular}
% \caption{FID and }
% \label{tab:clip}
% \end{table}

\begin{table}
\centering
\caption{\textbf{Quantitative Comparisons and User Study}. We compare our \modelname~with recent text-to-3D works~\cite{poole2022dreamfusion,wang2024prolificdreamer,yi2023-gaussiandreamer} in two different CLIP-scores~\cite{radford2021clip}. We also invest the human evaluation between the GaussianDreawmer and our \modelname.}
\renewcommand{\tabcolsep}{3mm}
\begin{tabular}{l|ccc}
\toprule
\textbf{Method} &  ViT-L/14 & ViT-bigG-14 & Human-evaluation \\\midrule
DreamFusion & 0.40 & 2.37 & - \\
ProlificDreamer & 2.93 & 9.23 & - \\
GaussianDreamer & 5.06 & 17.21 & 32\% \\ \midrule
\modelname & \textbf{5.64} & \textbf{18.50} & \textbf{68\%}\\
\bottomrule
\end{tabular}
\label{tab0:quantitative}
\end{table}

% 1. Clip score experiment
% 2. User study
% 3. 3D数据集evaluation

\begin{figure}
\centering   
  \includegraphics[width=0.5\textwidth]{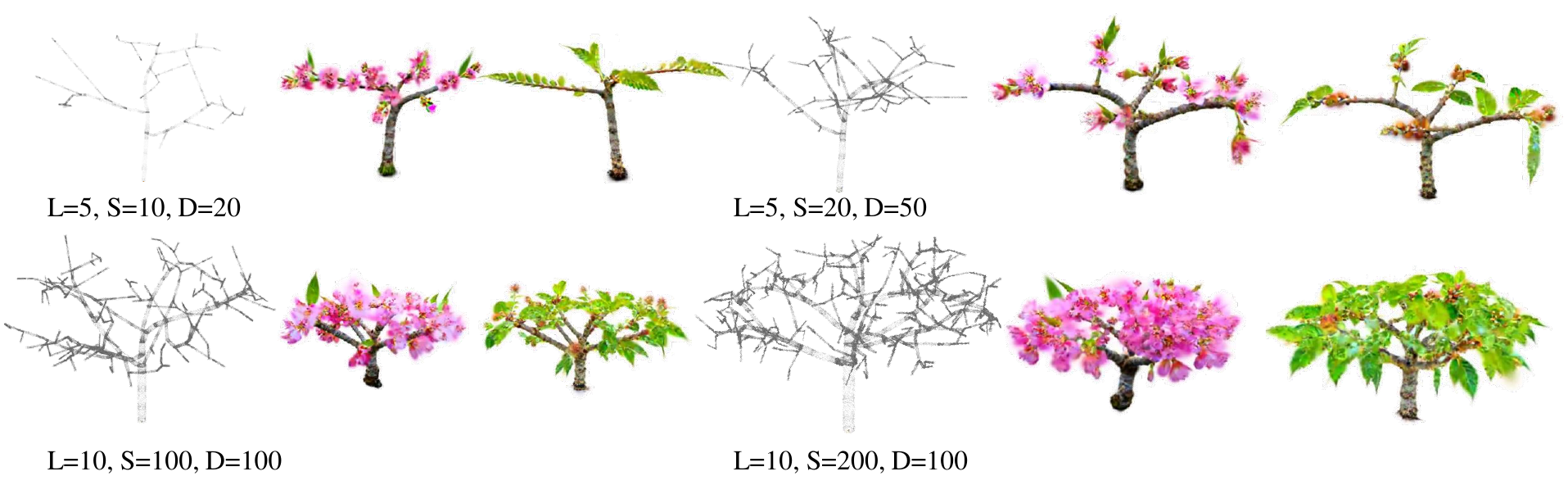}
\caption{\textbf{Ablation study of the various complexity of structures}. "L" denotes segment length, "S" represents the radius of the initial point cloud range, and "D" indicates the initial point cloud density. }
    \label{fig3:structure}
\vspace{-10pt}
\end{figure}

% 1. 点云对比图片 2. 6-9个其他prompt不同风格的图片
% 1. Qualitative comparisons, 对比其他text-to-image的效果，用A tree/A bonsai对比。
% 2. More generation cases with two views 3X3=9 cases，大概贴9个不同的case
% 3. 3D prior comparison Shape/pointE/ours
\subsection{Ablation study and Analysis}
 \textbf{Structures of varying complexity.}
We also provided two different cases guided by structures of varying complexity to demonstrate the effectiveness of our fine structure 3DGS under different complex structure conditions, as shown in Figure ~\ref{fig3:structure}. "L" denotes segment length, "S" represents the radius of the initial point cloud range, and "D" indicates the initial point cloud density. Higher values of these parameters result in more complex final structures. 
Our ablation study indicates that details such as flowers and leaves are rendered more clearly when the complexity of the structure is lower. 
However, with increasing structural complexity, detailed features such as flowers and foliage may exhibit noise or reduced quality, despite branch structures still complying with the 3D structural prior. This issue arises due to the rendering process utilizing only four randomly selected camera angles, which are likely to be obstructed by complex structures, thereby introducing noise into the rendered results.

 \textbf{Analysis on fine structure and coarse structure pipelines.}
The results of fine structure and coarse structure 3DGS are compared in Figure \ref{fig1:result}, demonstrating distinct outcomes for each approach. The fine structure method allows the generated 3D Gaussian bonsai to closely match the input structure, precisely fitting complex structural guides. On the other hand, the coarse structure method aligns more closely with the semantics from the prompt and showcases a greater diversity in bonsai. Both methods exhibit significantly higher quality compared to the baseline. 
Another analysis focuses on why multi-view structure consistency learning cannot be incorporated into the fine structure approach. The constraint imposed by the 3D structure prior in fine structure 3DGS is quite strong, resulting in a very clear structure in the diffusion rendering process. Adding multi-view structure consistency learning to this method would not enhance the definition of the structure further. Instead, the inherent randomness of SD may introduce additional noise, reducing the overall performance and clarity.

\section{Conclusion and Future Work}
In this paper, we propose a novel framework for generating structure-aware bonsai models using conditioned 3D Gaussian splatting. Our framework facilitates the generation of high-quality bonsai even under complex structural conditions. The major limitation is that our model may exhibit noise and blurriness when generating objects with extremely complex structures. This issue arises because the existing 3D framework relies on multi-view 2D diffusion, which typically features a limited number of camera angles. This work provides a robust foundation for complex 3D asset generation, stimulating critical analysis of current framework limitations and enhancement opportunities for realistic 3D generation.

\bibliographystyle{IEEEbib}
\bibliography{icme2025references}

\end{document}

% \title{Supplement}

% \author{Anonymous ICME submission}

% \maketitle

\section{Supplementary}
\subsection{Preliminary}

\textbf{3D Gaussian Splatting} \textit{3D Gaussian Splatting (3DGS)} is an innovative method for creating new viewpoints. Unlike implicit neural representation (INR) techniques like NeRF, 3DGS renders images through splatting, enabling real-time performance. Specifically, 3DGS models the scene using a collection of anisotropic Gaussians, each defined by its center position $\mu \in \mathbb{R}^3$, covariance $\Sigma \in \mathbb{R}^7$, color $c \in \mathbb{R}^3$, and opacity $\alpha \in \mathbb{R}^1$. And the 3D Gaussians can be queried as follows:
\begin{equation}
G(x) = e^{-\frac{1}{2}(x)^T\Sigma^{-1}(x)},
\end{equation}
where $x$ represents the distance between $\mu$ and the query point. A typical neural point-based rendering approach is employed to compute the color of each pixel. A ray $r$ is cast from the camera's center, and the color and density of the 3D Gaussians intersected by the ray are computed along its path. The rendering process is described by:
\begin{equation}
C(r) = \sum_{i \in N} c_i \prod_{j=1}^{i-1} (1 - \sigma_j), \quad \sigma_i = \alpha G(x_i),
\end{equation}
where \( C \) represents the set of all Gaussians intersected by the ray \( r \), and \( c_i \) is the color of the \( i \)-th Gaussian.

\textbf{Shap-E} \textit{Shap-E} is a conditional generative model for 3D assets that leverages prior information from point clouds and images. By combining with Neural Radiance Fields (NeRF), Shap-E achieves a precise 3D representation. This methodology employs a 3D encoder and a NeRF-based decoder, facilitating the joint training of neural networks to acquire accurate 3D representations. In Shap-E, the 3D encoder utilizes the information from point clouds and integrates it with patch-based image data. Through cross-attention mechanisms, the encoder can obtain latent projections. The encoder outputs parameters required for latent rendering, including non-negative density value $\sigma$, RGB information, and Signed Distance Functions (SDF) information. This enables the MLP-based NeRF to decode information from point clouds accurately, achieving an appropriate 3D representation. Shap-E employs a joint loss function composed of $L_{RGB}$ loss, $L_{T}$ loss, and $L_{STF}$ loss for fine-tuning. The loss function is formalized as follows:
\begin{equation}
L_{FT} = L_{RGB} + L_{T}+ L_{STF}
\end{equation}

This approach efficiently leverages point clouds as prior information, significantly enhancing the quality of generation. In fact, this approach based on 3D priors can be effectively integrated with methods based on 3D Gaussian Splatting, markedly improving both the quality and speed of rendering.

\begin{algorithm}
\caption{3D Space Colonization Algorithm with Learnable Parameters\\$A$: Set of attractor points, representing the targets for growth.\\$B$: Set of branch nodes, starting from the root node.\\$\Delta l$: Growth step, the distance each branch grows in one step.\\$d_{\text{kill}}$: Kill distance, the threshold distance within which attractor points are removed.\\$d_{\text{influence}}$: Influence distance, the range within which attractor points influence branch growth.\\$\theta$: Learnable parameters, which adjust the influence weight of each attractor point.\\$\text{weight}(\mathbf{v}, \theta)$: A function that computes the influence weight based on the vector $\mathbf{v}$ and learnable parameters $\theta$.}
\begin{algorithmic}[1]
\State \textbf{Initialize:} attractors $A = \{a_1, a_2, \ldots, a_m\}$, tree branches $B = \{b_1\}$ , growth step $\Delta l$, kill distance $d_{\text{kill}}$, influence distance $d_{\text{influence}}$, learnable parameters $\theta = \{\theta_1, \theta_2, \ldots, \theta_k\}$

\While{not all attractors removed or growth limit reached}
    \State Initialize new branches set $B_{\text{new}} = \{\}$
    \For{each branch $b$ in $B$}
        \State Initialize growth direction $\mathbf{d} = (0, 0, 0)$
        \State Initialize attractor count $count = 0$
        \For{each attractor $a$ in $A$}
            \State Compute vector $\mathbf{v} = a - b$
            \State Compute distance $dist = \|\mathbf{v}\|$
            \If{$dist < d_{\text{kill}}$}
                \State Remove $a$ from $A$
            \ElsIf{$dist < d_{\text{influence}}$}
                \State Compute weighted influence $w = \text{weight}(\mathbf{v}, \theta)$
                \State $\mathbf{d} \mathrel{+}= w \cdot \frac{\mathbf{v}}{dist}$
                \State $count \mathrel{+}= 1$
            \EndIf
        \EndFor
        \If{$count > 0$}
            \State $\mathbf{d} \mathrel{/}= count$
            \State Compute new branch position $b_{\text{new}} = b + \Delta l \cdot \mathbf{d}$
            \State Add $b_{\text{new}}$ to $B_{\text{new}}$
        \EndIf
    \EndFor
    \State Add $B_{\text{new}}$ to $B$
    \State Update learnable parameters $\theta$ using actual plant data
\EndWhile
\end{algorithmic}
\end{algorithm}

\begin{figure*}
\centering   
  \includegraphics[width=1\textwidth]{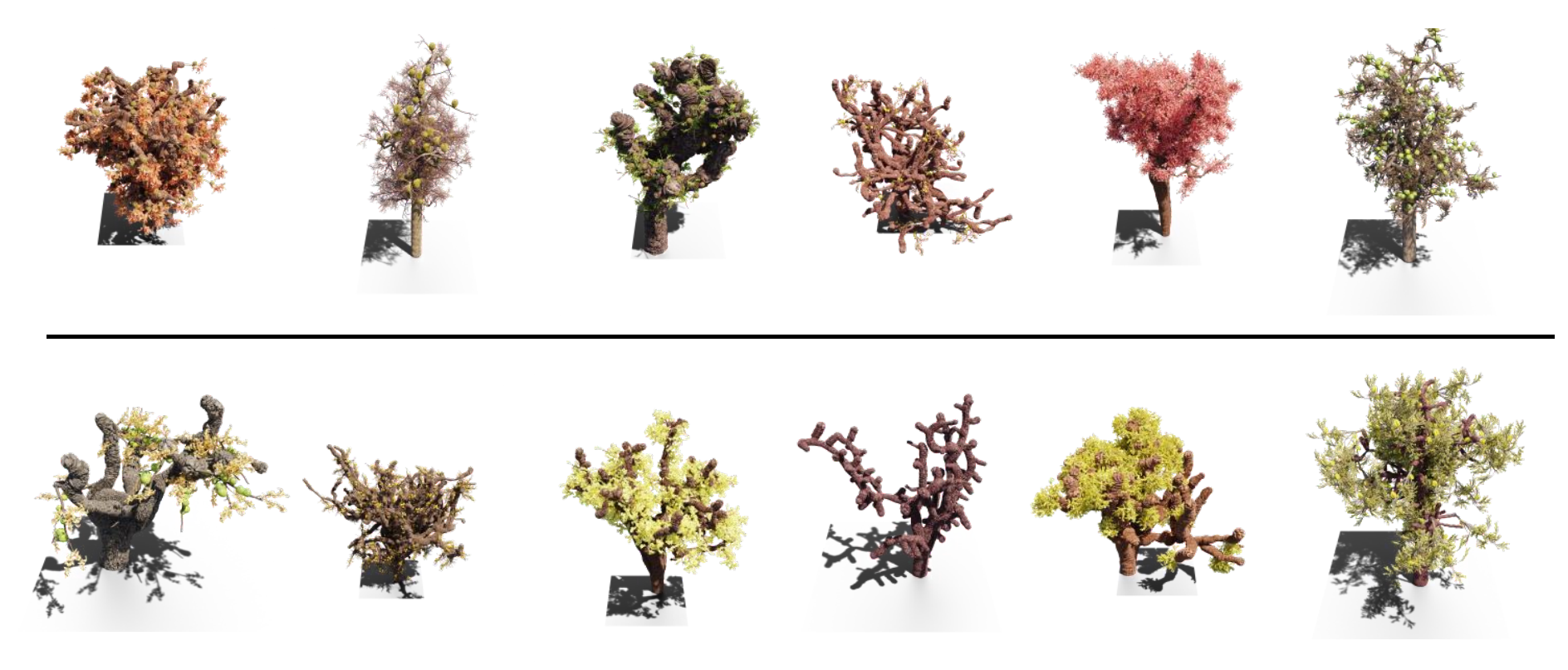}
  \vspace{-10pt}
\caption{3D tree assets generated by Infinigen~\cite{raistrick2023-infinigen}}
    \label{fig:infinigen}
    \vspace{-10pt}
\end{figure*}

\begin{figure*}
\centering   
  \includegraphics[width=1\textwidth]{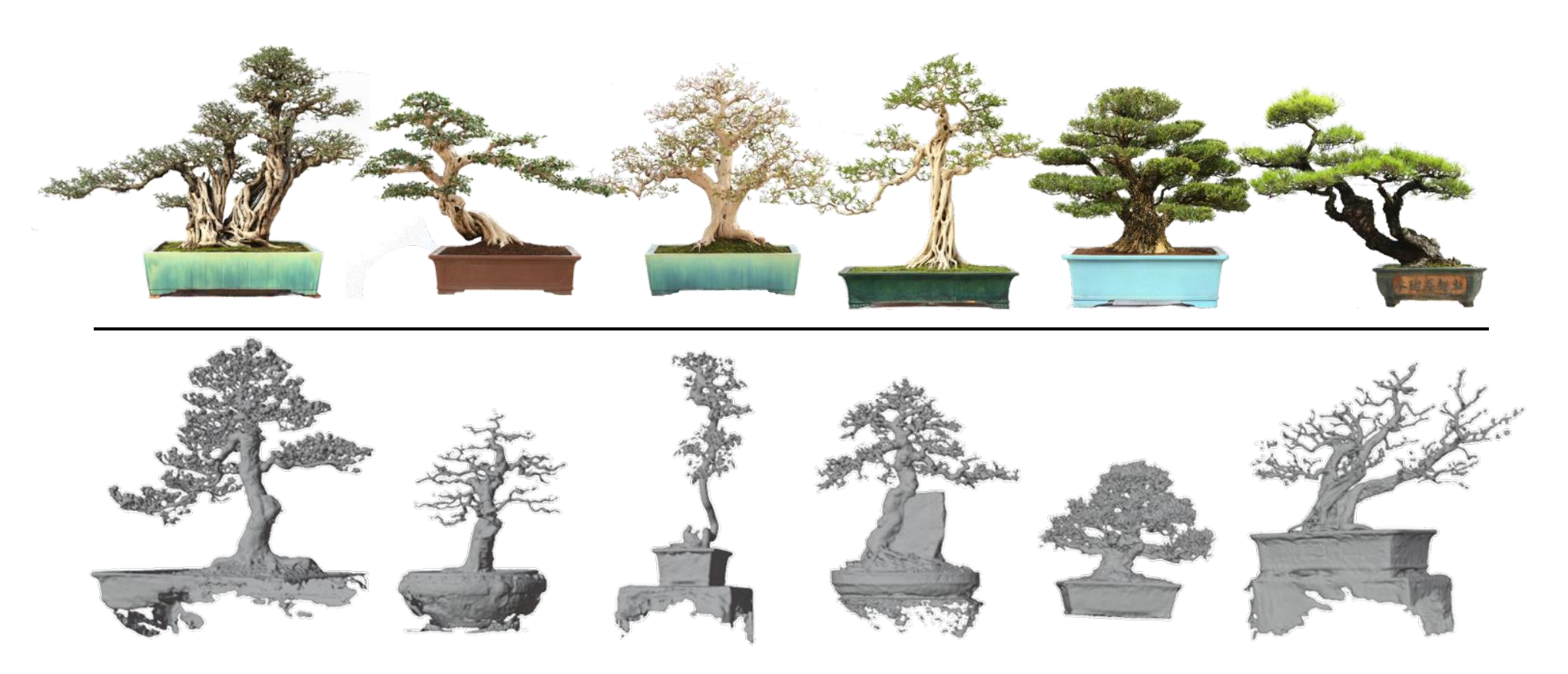}
  \vspace{-10pt}
\caption{Some cases from our 2D and 3D bonsai dataset.}
    \label{fig:dataset}
    \vspace{-10pt}
\end{figure*}

\begin{figure*}
\centering   
  \includegraphics[width=1\textwidth]{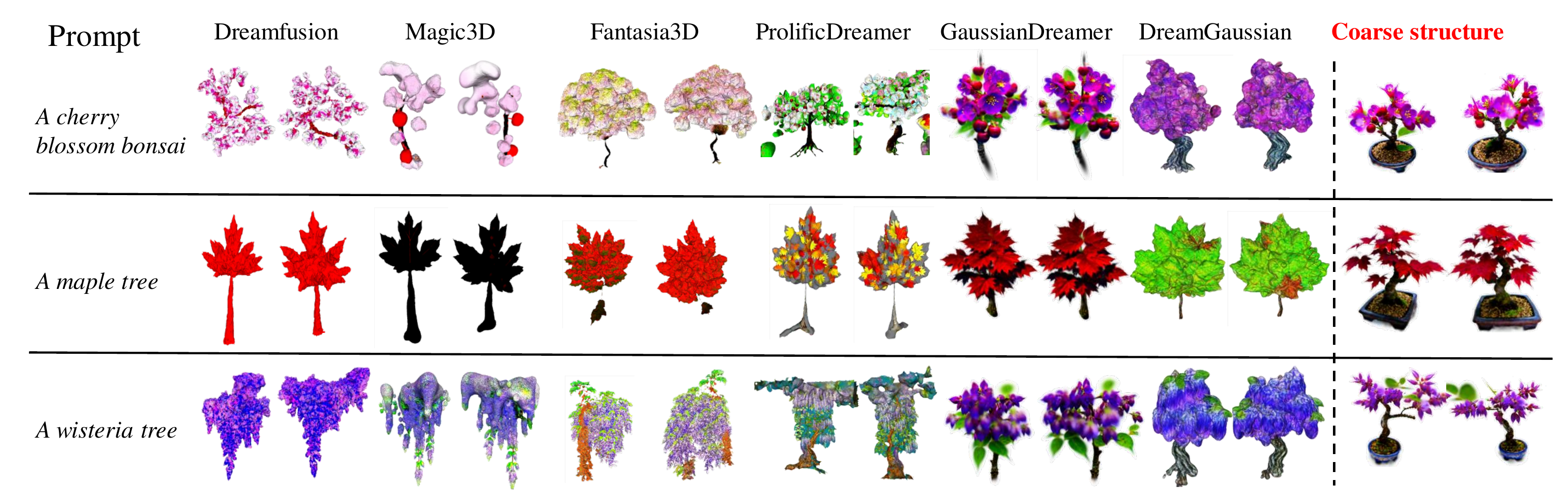}
  \vspace{-10pt}
\caption{More qualitative comparisons for image-to-3D evaluations between our \modelname(coarse structure)~and DreamFusion~\cite{poole2022dreamfusion}, Magic3D~\cite{lin2023magic3d}, Fantasia3D~\cite{chen2023fantasia3d}, ProlificDreamer~\cite{wang2024prolificdreamer}, GaussianDreamer~\cite{yi2023-gaussiandreamer}, DreamGaussian~\cite{tang2023-dreamgaussian}.}
    \label{fig:supp_result}
    \vspace{-10pt}
\end{figure*}

\subsection{3D Space Colonization Algorithm}
Space Colonization Algorithm(SCA) is commonly applied in generating complex structures, and it is especially widely used in procedural tree generation methods. Trees generated via SCA closely resemble real trees, making them suitable for various tree image and point cloud generation tasks.

The fundamental idea is outlined as follows:
The operation of the algorithm begins with an initial configuration of \( N \) attraction points (usually hundreds or thousands) and one or several tree nodes. The tree is generated iteratively. In each iteration, an attraction point may influence the tree node that is closest to it. This influence occurs if the distance between the point and the closest node is less then a radius of influence \( d_i \). There may be several attraction points that influence a single tree node \( v \): we denote this set of points by \( S(v) \). If \( S(v) \) is not empty, a new tree node \( v' \) will be created and attached to \( v \) by segment \( (v, v') \). The node \( v' \) is positioned at a distance \( D \) from \( v \), in the direction defined as the average of the normalized vectors toward all the sources \( s \in S(v) \). Thus, \( v' = v + D\hat{n} \), where

\begin{equation}
\hat{n} = \frac{\tilde{n}}{\|\tilde{n}\|} \quad \text{and} \quad \tilde{n} = \sum_{s\in S(v)} \frac{s - v}{\|s - v\|}.
\end{equation}

Optionally, the direction of growth can be biased by a vector \( \tilde{g} \) representing a combined effect of branch weight and tropisms using the equation
\begin{equation}
\tilde{n} = \frac{\hat{n} + \tilde{g}}{\|\hat{n} + \tilde{g}\|}.
\end{equation}

Once the new nodes have been added, a check is performed to test which, if any, of the attraction points should be removed due to the proximity of tree branches that have grown toward these points.

We introduce an improved trainable space colonization algorithm extended to 3D space, which originally applied in two dimensions, to adapt to 3D model generation. Our method utilizes it to generate the skeletal structure of trees as 3D prior information for the 3D generation pipeline. This multi-view synthesis approach based on 3D prior information aids in maintaining 3D consistency to optimize the consistency and rendering quality of 3D models from multiple perspectives. More importantly, this structure-aware method shows immense potential in the 3D generation of complex objects. It maintains 3D consistency in 3D representation through a fixed skeletal structure, adapting to the structural generation of complex objects, thereby improving issues like Multi-face and distortions. Algorithm 1 illustrates the detailed flowchart of the 3D SCA.

\subsection{Some Result from Infinigen}
We utilized Infinigen~\cite{raistrick2023-infinigen} to generate examples based on Blender methods, with these assets relying heavily on resources in Blender. Consequently, their diversity is quite limited, and it is evident that the cases generated do not accurately reflect the natural distribution of plant branches.

\bibliographystyle{IEEEbib}
\bibliography{icme2025references}